%% file: main.tex
\documentclass[final]{cvpr}

\usepackage{times}
\usepackage{epsfig}
\usepackage{graphicx}
\usepackage{amsmath}
\usepackage{amssymb}
\newcommand*{\rom}[1]{\expandafter\@slowromancap\romannumeral #1@}

\usepackage[pagebackref=true,breaklinks=true,colorlinks,bookmarks=false]{hyperref}

\def\cvprPaperID{1407} %
\def\confYear{CVPR 2021}

\usepackage{mathrsfs}
\usepackage{amsmath}
\DeclareMathOperator*{\argmax}{arg\,max}
\DeclareMathOperator*{\argmin}{arg\,min}

\usepackage{amssymb}
\usepackage{amsthm}
\usepackage{amscd}
\usepackage{amsfonts,dsfont}
\usepackage{graphicx}
\usepackage{algpseudocode}
\usepackage[ruled,vlined]{algorithm2e}
\usepackage{enumitem}

\usepackage{threeparttable}
\usepackage{multirow}

\newcommand{\estD}{{\hat D}}
\newcommand{\estDg}{{\hat D^{\tau}}}
\newcommand{\estDisg}{{\hat {\mathcal D}^{\tau}}}
\newcommand{\estDis}{{\hat {\mathcal D}}}

\newcommand{\PP}{\mathbb P}
\newcommand{\BR}{\mathds 1}

\newcommand{\E}{\mathbb E}

\newcommand{\squishlist}{
\begin{list}{{{\small{$\bullet$}}}}
{\setlength{\itemsep}{3pt}      \setlength{\parsep}{1pt}
\setlength{\topsep}{1pt}       \setlength{\partopsep}{0pt}
\setlength{\leftmargin}{1em} \setlength{\labelwidth}{1em}
\setlength{\labelsep}{0.5em} } }
\newcommand{\squishend}{  \end{list}  }
\newcommand{\SPL}{CORES$^2$}

\def\BState{\State\hskip-\ALG@thistlm}

\newtheorem{theorem}{Theorem}
\newtheorem{corollary}{Corollary}

\newtheorem{lemma}{Lemma}

\usepackage{hyperref}

\ifodd 0
\newcommand{\rev}[1]{{\color{blue}#1}}

\newcommand{\clar}[1]{\textbf{\color{green}(NEED CLARIFICATION: #1)}}
\newcommand{\response}[1]{\textbf{\color{magenta}(RESPONSE: #1)}}

\else
\newcommand{\rev}[1]{#1}
\newcommand{\clar}[1]{}
\newcommand{\response}[1]{}
\fi
\newcommand{\RNum}[1]{\uppercase\expandafter{\romannumeral #1\relax}}

\newcommand{\covpeer}{\text{CAL}}
\usepackage{pifont}%
\newcommand{\cmark}{\ding{51}}%
\newcommand{\xmark}{\ding{55}}%

\newcommand{\algcom}[1]{\textsl{\color{blue}{\footnotesize #1}}}

\usepackage[utf8]{inputenc} %
\usepackage[T1]{fontenc}    %
\begin{document}

\title{A Second-Order Approach to Learning with Instance-Dependent Label Noise}

\author{Zhaowei Zhu$^{\dagger}$ \quad Tongliang Liu$^{\mathsection}$ \quad  Yang Liu$^\dagger$\\
$^\dagger$Computer Science and Engineering, University of California, Santa Cruz \\ $^\mathsection$Trustworthy Machine Learning Lab, The University of Sydney\\
{\tt\small $^{\dagger}$\{zwzhu,yangliu\}@ucsc.edu}
\quad
{\tt\small $^{\mathsection}$tongliang.liu@sydney.edu.au}
}

\maketitle
\pagestyle{empty}  %
\thispagestyle{empty} %

\begin{abstract}
The presence of label noise often misleads the training of deep neural networks. Departing from the recent literature which largely assumes the label noise rate is only determined by the true label class, the errors in human-annotated labels are more likely to be dependent on the difficulty levels of tasks, resulting in settings with instance-dependent label noise. We first provide evidences that the heterogeneous instance-dependent label noise is effectively down-weighting the examples with higher noise rates in a non-uniform way and thus causes imbalances, rendering the strategy of directly applying methods for class-dependent label noise questionable. Built on a recent work peer loss \cite{liu2019peer}, we then propose and study the potentials of a second-order approach that leverages the estimation of several covariance terms defined between the instance-dependent noise rates and the Bayes optimal label. We show that this set of second-order statistics successfully captures the induced imbalances. We further proceed to show that with the help of the estimated second-order statistics, we identify a new loss function whose expected risk of a classifier under instance-dependent label noise is equivalent to a new problem with only class-dependent label noise. This fact allows us to apply existing solutions to handle this better-studied setting. We provide an efficient procedure to estimate these second-order statistics without accessing either ground truth labels or prior knowledge of the noise rates. Experiments on CIFAR10 and CIFAR100 with synthetic instance-dependent label noise and Clothing1M with real-world human label noise verify our approach. Our implementation is available at \url{https://github.com/UCSC-REAL/CAL}.
\end{abstract}

\input{src/intro}

\input{src/pre}
\input{src/formu_first}

\input{src/formu_second}

\input{src/exp}

\input{src/conclusion}

\clearpage
\newpage
\bibliographystyle{ieee_fullname}
\bibliography{main.bib}

\input{src/appendix}

\end{document}

%% file: src/intro.tex
\vspace{-10pt}
\section{Introduction}
Deep neural networks (DNNs) are powerful in revealing and fitting the relationship between feature $X$ and label $Y$ when a sufficiently large dataset is given. However, the label $Y$ usually requires costly human efforts for accurate annotations. With limited budgets/efforts, the resulting dataset would be noisy, and the existence of label noise may mislead DNNs to learn or memorize wrong correlations \cite{han2020sigua,han2020survey,wang2021learning,xia2021robust,zhang2016understanding}.
To make it worse, the label noise embedded in human annotations is often instance-dependent, e.g., some difficult examples are more prone to be mislabeled \cite{wang2021tackling}. This hidden and imbalanced distribution of noise often has a detrimental effect on the training outcome \cite{jiang2020beyond,liu2021importance}. %
It remains an important and challenging task to learn with instance-dependent label noise. %

Theory-supported works addressing instance-dependent label noise mostly rely on loss correction, which requires estimating noise rates \cite{xia2020parts}. Recent work has also considered the possibility of removing the dependency on estimating noise rates \cite{sieve2020}. The proposed solution uses a properly specified regularizer to eliminate the effect of instance-dependent label noise. The common theme of the above methods is the focus on learning the underlying clean distribution by using certain forms of \underline{first-order statistics} of model predictions. In this paper, we propose a second-order approach with the assistance of additional \underline{second-order statistics} and explore how this information can improve the robustness of learning with instance-dependent label noise.
Our main contributions summarize as follows.

\begin{enumerate}[itemsep = -5pt, topsep = -7pt, leftmargin = 12pt]
    \item Departing from recent works \cite{sieve2020,liu2019peer,natarajan2013learning,patrini2017making,vahdat2017toward,xia2020parts,xiao2015learning} which primarily rely on the first-order statistics (i.e. expectation of the models' predictions) to improve the robustness of loss functions, we propose a novel second-order approach and emphasize the importance of using second-order statistics \rev{(i.e. several covariance terms)} when dealing with instance-dependent label noise. %
    \item 
    \rev{With the perfect knowledge of the covariance terms defined above, we identify a new loss  function that transforms the expected risk of a classifier under instance-dependent label noise to a risk with only class-dependent label noise, which is an easier case and can be handled well by existing solutions. Based on peer loss \cite{liu2019peer}, we further show the expected risk of class-dependent noise is equivalent to an affine transformation of the expected risk under the Bayes optimal distribution.
    Therefore we establish that our new loss function for \emph{Covariance-Assisted Learning (CAL)} will induce the same minimizer as if we can access the clean Bayes optimal labels.}
    \item  We show how the second-order statistics can be estimated efficiently using existing sample selection techniques. For a more realistic case where the covariance terms cannot be perfectly estimated, we prove the worst-case performance guarantee of our solution. 
    \item In addition to the theoretical guarantees, the performance of the proposed second-order approach is tested on the CIFAR10 and CIFAR100 datasets with synthetic instance-dependent label noise and the Clothing1M dataset with real-world human label noise. 
\end{enumerate}

\subsection{Related Works}

Below we review the most relevant literature. %

\noindent\textbf{Bounded loss functions}~~
Label noise encodes a different relation between features and labels. A line of literature treats the noisy labels as outliers.
However, the convex loss functions are shown to be prone to mistakes when outliers exist \cite{long2010random}.
To handle this setting, the cross-entropy (CE) loss can be generalized by introducing temperatures to  logarithm functions and exponential functions \cite{amid2019robust,amid2019two,zhang2018generalized}.
Noting the CE loss grows explosively when the prediction $f(x)$ approaches zero, some solutions focus on designing bounded loss functions \cite{ghosh2017robust,gong2018decomposition,shu2020learning,wang2019symmetric}.
These methods focus on the numerical property of loss functions, and most of them do not discuss the type of label noise under treatment.

\noindent\textbf{Learning clean distributions}~~
To be noise-tolerant \cite{manwani2013noise}, it is necessary to understand the effect of label noise statistically.
With the class-dependent assumption, the loss can be corrected/reweighted when the noise transition $T$ is available, which can be estimated by discovering anchor points \cite{liu2015classification,patrini2017making,xia2020extended}, exploiting clusterability \cite{zhu2021clusterability}, regularizing total variation \cite{zhang2021learning}, or minimizing volume of $T$ \cite{li2021provably}.
The loss correction/reweighting methods rely closely on the quality of the estimated noise transition matrix.
To make it more robust, an additive slack variable $\Delta T$ \cite{xia2019anchor} or a multiplicative dual $T$ \cite{dualT2020nips} can be used for revision.
Directly extending these loss correction methods to instance-dependent label noise is prohibitive since the transition matrix will become a function of feature $X$ and the number of parameters to be estimated is proportional to the number of training instances. 
Recent follow-up works often introduce extra assumption \cite{xia2020parts} or measure  \cite{berthon2020confidence}.
Statistically, the loss correction approach is learning the underlying clean distribution if a perfect $T$ is applied. 
When the class-dependent noise rate is known, surrogate loss \cite{natarajan2013learning}, an unbiased loss function targeting on binary classifications, also learns the clean distribution.
Additionally, the symmetric cross-entropy loss \cite{wang2019symmetric}, an information-based loss $L_{\sf DMI}$ \cite{xu2019l_dmi}, a correlated agreement (CA) based loss peer loss \cite{liu2019peer}, and its adaptation for encouraging confident predictions \cite{sieve2020} are proposed to learn the underlying clean distribution without knowing the noise transition matrix.

\noindent\textbf{Other popular methods}~~
Other methods exist with more sophisticated training framework or pipeline, including sample selection \cite{sieve2020,han2018co,jiang2017mentornet,lee2018cleannet,wei2020combating,yu2019does,yao2020searching}, label correction \cite{han2019deep,li2017learning,veit2017learning}, and semi-supervised learning \cite{Li2020DivideMix,nguyen2019self}, etc.

%% file: src/pre.tex
\section{Preliminaries}\label{Sec:pre}

This paper targets on a classification problem given a set of $N$ training examples with Instance-Dependent label Noise (IDN) denoted by $\widetilde D:=\{( x_n,\tilde y_n)\}_{n\in  [N]}$, where $[N] := \{1,2,\cdots,N\}$ is the set of indices.
The corresponding noisy data distribution is denoted by $\widetilde{\mathcal D}$.
Examples $(x_n,\tilde y_n)$ are drawn according to random variables $(X,\widetilde Y) \sim \widetilde{\mathcal D}$. 
Our goal is to design a learning mechanism that is guaranteed to be robust when learning with only accessing $\widetilde D$.
Before proceeding, we summarize important definitions as follows.

\noindent\textbf{Clean distribution $\mathcal D$}~~
Each noisy example $(x_n,\tilde y_n)\in\widetilde D$ corresponds to a clean example $(x_n,y_n) \in D$, which contains one \emph{unobservable} ground-truth label, a.k.a. clean label.
Denote by $\mathcal D$ the clean distribution. Clean examples $(x_n,y_n)$ are drawn from random variables $(X,Y)\sim \mathcal D$.

\noindent\textbf{Bayes optimal distribution $\mathcal D^*$}~~
Denote by $Y^*$ the Bayes optimal label given feature $X$, that is:
$
Y^*|X := \argmax_Y ~\PP(Y|X), (X,Y)\sim \mathcal D.
$
The distribution of $(X,Y^*)$ is denoted by $\mathcal D^*$. Note the Bayes optimal distribution $\mathcal D^*$ is different from the clean distribution $\mathcal D$ when $\PP(Y|X) \notin \{0,1\}$.
Due to the fact that the information encoded between features and labels is corrupted by label noise, and both clean labels and Bayes optimal labels are unobservable, inferring the Bayes optimal distribution $\mathcal D^*$ from the noisy dataset $\widetilde D$ is a non-trivial task.
\rev{Notably there exist two approaches \cite{sieve2020,cheng2017learningdistill} that provide guarantees on constructing the Bayes optimal dataset. %
We would like to remind the readers that the noisy label $\tilde y_n$, clean label $y_n$, and Bayes optimal label $y_n^*$ for the same feature $x_n$ may disagree with each other.}

Most of our developed approaches will focus on dealing with the Bayes optimal distribution $\mathcal D^*$. 
{By referring to $\mathcal D^*$, as we shall see later, we are allowed to estimate the second-order statistics defined w.r.t. $Y^*$.} 

\noindent\textbf{Noise transition matrix $T(X)$}~~
Traditionally, the noise transition matrix is defined based on the relationship between clean distributions and noisy distributions \cite{sieve2020,liu2019peer,Patrini_2017_CVPR,xia2020parts}.
In recent literature \cite{cheng2017learningdistill}, the Bayes optimal label (a.k.a. distilled label in \cite{cheng2017learningdistill}) also plays a significant role.
In the image classification tasks where the performance is measured by the clean test accuracy, predicting the Bayes optimal label achieves the best performance. 
This fact motivates us to define a new noise transition matrix based on the Bayes optimal label as follows:
\[
T_{i,j}(X) = \PP( \widetilde Y = j | Y^* = i,X),
\]
where $T_{i,j}(X)$ denotes the $(i,j)$-th element of the matrix $T(X)$. 
Its expectation is defined as $T:= \E [T(X)]$, with the $(i,j)$-th element being $T_{i,j}:= \E[T_{i,j}(X)]$.

\noindent\textbf{Other notations}~~
Let $\mathcal X$ and $\mathcal Y$\footnote{We focus on the closed-set label noise, i.e. $\widetilde Y$, $Y$, and $Y^*$ share the same space $\mathcal Y$.} be the space of feature $X$ and label $Y$, respectively. 
The classification task aims to identify a classifier $f: \mathcal X \rightarrow \mathcal Y$ that maps $X$ to $Y$ accurately.
One common approach is minimizing the empirical risk using DNNs with respect to the \emph{cross-entropy (CE) loss} defined as:
$
    \ell(f(X),Y) := - \ln(f_X[Y]), ~ Y \in [K],
$
where $f_X[Y]$ denotes the $Y$-th component of $f(X)$ and $K$ is the number of classes.
Let $\BR\{\cdot\}$ be the indicator function taking value $1$ when the specified condition is satisfied and $0$ otherwise. Define the \emph{0-1 loss} as 
$\BR{(f(X),Y)}:=\BR\{f(X) \ne Y\}.$
Define the Bayes optimal classifier $f^*$ as 
$
f^* = \argmin_{f} ~\E_{\mathcal D^*}[\BR(f(X),Y^*)].
$
Noting the CE loss is classification-calibrated \cite{bartlett2006convexity}, %
given enough clean data, the Bayes optimal classifier can be learned using the CE loss: 
$
f^* \hspace{-2pt}=\hspace{-2pt} \argmin_{f} ~\E_{\mathcal D}[\ell(f(X),Y)].
$

\noindent\textbf{Goal}~~
Different from the goals in surrogate loss \cite{natarajan2013learning}, $L_{\sf{DMI}}$ \cite{xu2019l_dmi}, peer loss \cite{liu2019peer}, and CORES$^2$ \cite{sieve2020}, which focus on recovering the performance of learning on clean distributions, we aim to learn a classifier $f$ from the noisy distribution $\widetilde{\mathcal D}$ which also minimizes $\E [\BR(f(X), Y^*)], (X,Y^*)\sim \mathcal D^*$. 
Note $\E[\BR(f^*(X), Y^*)] = 0$ holds for the Bayes optimal classifier $f^*$. Thus, in the sense of searching for the Bayes optimal classifier, our goals are aligned with the ones focusing on the clean distribution. 

%% file: src/formu_first.tex
\section{Insufficiency of First-Order Statistics}\label{sec:cal}
Peer loss \cite{liu2019peer} and its inspired confidence regularizer \cite{sieve2020} are two recently introduced robust losses that operate without the knowledge of noise transition matrices, which presents them as preferred solutions for more complex noise settings. 
In this section, we will first review the usages of first-order statistics in peer loss and the confidence regularizer (Section~\ref{sec:first-order-info}), and then analyze the insufficiency of using only the first-order statistics when handling the challenging IDN (Section~\ref{sec:pure_peer_IDN}).
Besides, we will anatomize the down-weighting effect of IDN and provide intuitions for how to make IDN easier to handle (Section~\ref{sec:down-weight}).

 We formalize our arguments using peer loss, primarily due to 1) its clean analytical form, and 2) that our later proposed solution will be built on peer loss too. Despite the focus on peer loss, we believe these observations are generally true when other existing training approaches meet IDN. 

For ease of presentation, the following analyses focus on binary cases (with classes $\{-1,+1\}$).
Note the class $-1$ should be mapped to class $0$ following the notations in Section~\ref{Sec:pre}. 
For a clear comparison with previous works, we follow the notation in \cite{liu2019peer} and use class $\{-1,+1\}$ to represent classes $\{0,1\}$ when $K=2$.
The error rates in $\widetilde Y$ are then denoted as $e_+(X):=\PP(\widetilde Y = -1|Y^*=+1,X)$, $e_-(X):=\PP(\widetilde Y = +1|Y^*=-1,X)$. Most of the discussions generalize to the multi-class setting. 
 
\subsection{Using First-Order Statistics in Peer Loss}\label{sec:first-order-info}

It has been proposed and proved in peer loss \cite{liu2019peer} and CORES$^2$ \cite{sieve2020} that the learning could be robust to label noise by considering some first-order statistics related to the model predictions.
For each example $(x_n,\tilde y_n)$, peer loss \cite{liu2019peer} has the following form:
\[
\ell_{\text{PL}}(f(x_n),\tilde y_{n}) := \ell(f(x_n),\tilde y_n) - \ell(f(x_{n_1}),\tilde y_{n_2}),
\]
where $(x_{n_1},\tilde y_{n_1})$ and $(x_{n_2},\tilde y_{n_2})$ are two randomly sampled peer samples for $n$.
The first-order statistics related to model predictions characterized by the peer term $ \ell(f(x_{n_1}),\tilde y_{n_2})$ are further extended to a confidence regularizer in CORES$^2$ \cite{sieve2020}:
\[
  \ell_{\text{CORES}^2}(f(x_n),\tilde y_{n}) :=    \ell(f(x_n),\tilde y_{n}) - \beta \E_{\mathcal{D}_{\widetilde{Y}|\widetilde{D}}}[\ell(f(x_n),\widetilde Y)],
\]
where $\beta$ is a hyperparameter controlling the ability of regularizer, and {\small $\mathcal{D}_{\widetilde{Y}|\widetilde{D}}$} is the marginal distribution of {\small $\widetilde Y$} given dataset {\small $\widetilde D$}.
Although it has been shown in \cite{sieve2020} that learning with an appropriate $\beta$ would be robust to instance-dependent label noise theoretically, in real experiments, converging to the guaranteed optimum by solving a highly non-convex problem is difficult.

\subsection{Peer Loss with IDN}\label{sec:pure_peer_IDN}

Now we analyze the possible performance degradation of using the binary peer loss function proposed in \cite{liu2019peer} to handle IDN. 
Denote by $$\tilde f^*_{\text{peer}}  := \argmin_{f} \E_{\widetilde {\mathcal D}} \left[ {\BR_{\text{PL}}}(f(X), \widetilde Y)  \right]$$ the optimal classifier learned by minimizing 0-1 peer loss, where ${\BR_{\text{PL}}}$ represents $\ell_{\text{PL}}$ with 0-1 loss (could also be generalized for $\ell_{\text{CORES}^2}$ with 0-1 loss).
\rev{Let $p^*:= \PP(Y^*=+1)$.}
With a bounded variance in the error rates, supposing 
\[
\E|e_+(X)-\E[e_+(X)]|\leq \epsilon_+, ~ \E|e_-(X)-\E[e_-(X)]| \leq \epsilon_-,
\]
the worst-case performance bound for using pure peer loss is provided in Theorem~\ref{thm:peerIDN} and proved in Appendix~\ref{proof:peerIDN}.

\begin{theorem}[Performance of peer loss]\label{thm:peerIDN}
With the peer loss function proposed in \cite{liu2019peer}, we have
\[
\E[\BR(\tilde f^*_{peer}(X),Y^*)] \leq \frac{2(\epsilon_+ + \epsilon_-)}{1-e_+-e_-} + 2|p^*-0.5|.
\]
\end{theorem}
Theorem~\ref{thm:peerIDN} shows the ratio of wrong predictions given by $\tilde f^*_{\text{peer}}$ includes two components. The former term $\frac{2(\epsilon_+ + \epsilon_-)}{1-e_+-e_-}$ is directly caused by IDN, indicating the error is increasing when the instance-dependent noise rates have larger mean (larger $e_+ + e_-$) and larger \rev{variation} (larger $\epsilon_+ + \epsilon_-$).
The latter term $2|p^*-0.5|$ shows possible errors induced by an unbalanced $\mathcal D^*$.
Theorem~\ref{thm:peerIDN} generalizes peer loss where $\epsilon_+=\epsilon_-=0$, i.e., the error rates are homogeneous across data instances, and there is no need to consider any second-order statistics that involve the distribution of noise rates.

\subsection{Down-weighting Effect of IDN}\label{sec:down-weight}

\rev{We further discuss motivations and intuitions by studying}
how IDN affects the training differently from the class-dependent one. Intuitively, a high noise rate reduces the informativeness of a particular example $(x,y)$, therefore ``down-weighting" its contribution to training. We now analytically show this under peer loss. %

As a building block, the invariant property (in terms of the \textit{clean distribution $\mathcal D$}) originally discovered by peer loss on class-dependent label noise is first adapted for the \textit{Bayes optimal distribution $\mathcal D^*$}. Define $e_- := \PP(\widetilde Y=+1|Y^* = -1)$ and $e_+ := \PP(\widetilde Y=-1|Y^* = +1)$.
Focusing on a particular class-dependent $\widetilde{\mathcal D}$, we provide Lemma~\ref{lem:invariant_d} and its proof in Appendix~\ref{proof:invariant_d}.
\begin{lemma}[Invariant property of peer loss \cite{liu2019peer}]\label{lem:invariant_d}
Peer loss is invariant to class-dependent label noise:
{\small \begin{equation}\label{eq:peerBinary}
    \E_{\widetilde{\mathcal D}}[{\BR_{\text{PL}}}(f(X),\widetilde{Y})] =(1-e_+-e_-)\E_{\mathcal D^*}[{\BR_{\text{PL}}}(f(X),Y^*)].
\end{equation}}
\end{lemma}

Then we discuss the effect of IDN.
Without loss of generality, consider a case where noisy examples are drawn from two noisy distributions $\widetilde {\mathcal D}_{\text{I}}$ and $\widetilde {\mathcal D}_{\text{II}}$, and the noise rate of $\widetilde {\mathcal D}_{\text{II}}$ is higher than $\widetilde {\mathcal D}_{\text{I}}$, i.e. $e_{+,\text{II}}+e_{-,\text{II}} > e_{+,\text{I}}+e_{-,\text{I}}$, where $e_{+,\text{I (II)}} := \PP_{\widetilde {\mathcal D}_{\text{I (II)}}}(\widetilde Y=-1|Y^* = +1)$. %
\rev{Assume a particular setting of IDN that the noise is class-dependent (but not instance-dependent) only within each distribution, and different between two distributions, i.e. \textit{part-dependent} \cite{xia2020parts}.}
Let ${\mathcal D}_{\text{I}}^*$ and ${\mathcal D}_{\text{II}}^*$ be the Bayes optimal distribution related to $\widetilde{\mathcal D}_{\text{I}}$ and $\widetilde{\mathcal D}_{\text{II}}$. 
For simplicity, we write $\PP((X,Y^*)\sim{\mathcal D}^*_{\text{I} (\text{II})}|(X,Y^*)\sim{\mathcal D}^*)$ as $\PP({\mathcal D}_{\text{I} (\text{II})}^*)$.
Then $\PP({\mathcal D}_{\text{I}}^*) = \PP(\widetilde{\mathcal D}_{\text{I}})$ and $\PP({\mathcal D}_{\text{II}}^*) = \PP(\widetilde{\mathcal D}_{\text{II}})$.
Note $\PP(\widetilde{\mathcal D}_{\text{I}}) e_{+,\text{I}} + \PP(\widetilde{\mathcal D}_{\text{II}}) e_{+,\text{II}} = e_{+}$ and $\PP(\widetilde{\mathcal D}_{\text{I}}) e_{-,\text{I}} + \PP(\widetilde{\mathcal D}_{\text{II}}) e_{-,\text{II}} = e_{-}.$ %
Then we have the following equality:
{
\begin{align*}
  & \E_{\widetilde{\mathcal D}}[{\BR_{\text{PL}}}(f(X),\widetilde{Y})] \\
= & \PP(\widetilde{\mathcal D}_{\text{I}})(1-e_{+,\text{I}}-e_{-,\text{I}})\E_{{\mathcal D}_{\text{I}}^*}[ {\BR_{\text{PL}}}(f(X),Y^*)]  \\  
  & + \PP(\widetilde{\mathcal D}_{\text{II}})(1-e_{+,\text{II}}-e_{-,\text{II}})\E_{\mathcal D_{\text{II}}^*}[ {\BR_{\text{PL}}}(f(X),Y^*)] \\
= & (1-e_{+,\text{I}}-e_{-,\text{I}}) \bigg( \PP(\mathcal D^*_{\text{I}})\E_{\mathcal D_{\text{I}}^*}[ {\BR_{\text{PL}}}(f(X),Y^*)]  \\
& + \frac{1-e_{+,\text{II}}-e_{-,\text{II}}}{1-e_{+,\text{I}}-e_{-,\text{I}}}\PP(\mathcal D^*_{\text{II}})\E_{\mathcal D_{\text{II}}^*}[ {\BR_{\text{PL}}}(f(X),Y^*)] \bigg),
\end{align*}}
where \[\frac{1-e_{+,\text{II}}-e_{-,\text{II}}}{1-e_{+,\text{I}}-e_{-,\text{I}}} < 1\] indicates \emph{down-weighting} examples drawn from $\widetilde {\mathcal D}_{\text{II}}$ (compared to the class-dependent label noise).

\noindent\textbf{What can we learn from this observation?}~~
First, we show the peer loss is already down weighting the importance of the more noisy examples. However, simply dropping examples with potentially high-level noise might lead the classifier to learn a biased distribution. \rev{Moreover, subjectively confusing examples are more prone to be mislabeled and critical for accurate predictions \cite{wang2021tackling}, thus need to be carefully addressed.} Our second observation is that if we find a way to compensate for the ``imbalances'' caused by the down-weighting effects shown above, the challenging instance-dependent label noise could be transformed into a class-dependent one, which existing techniques can then handle. %
More specifically, the above result shows the down-weighting effect is characterized by $T(X)$, implying only using the first-order statistics of model predictions without considering the distributions of the noise transition matrix $T(X)$ is insufficient to capture the complexity of the learning task. 
However, accurately estimating $T(X)$ is prohibitive since the number of parameters to be estimated is almost at the order of $O(NK^2)$ -- recall $N$ is the number of training examples and $K$ is the number of classes.
Even though we can roughly estimate $T(X)$, applying element-wise correction relying on the estimated $T(X)$ may accumulate errors.
Therefore, to achieve the transformation from the instance-dependent to the easier class-dependent, we need to resort to other statistical properties of $T(X)$.

%% file: src/formu_second.tex
\section{Covariance-Assisted Learning (CAL)}\label{sec:theory}

From the analyses in Section~\ref{sec:down-weight}, we know the instance-dependent label noise will ``automatically'' assign different weights to examples with different noise rates, thus cause imbalances.
When the optimal solution does not change under such down-weighting effects, the first-order statistics based on peer loss \cite{sieve2020,liu2019peer} work well. 
However, for a more robust and general solution, using additional information to ``balance'' the effective weights of different examples is necessary.
Although the Bayes optimal distribution is not accessible in real experiments, we first assume its existence for theoretical analyses in the ideal case, then we will discuss the gap to this optimal solution when we can only use a proxy $\estD$ that can be constructed efficiently. 

\subsection{Extracting Covariance from IDN}\label{sec:extension}

Again consider an instance-dependent noisy distribution $\widetilde{\mathcal D}$ with binary classes where $\widetilde Y\in\{-1,+1\}$.
Define the following two random variables \rev{to facilitate analyses}:
\[
Z_1(X):= 1-e_+(X)-e_-(X), ~Z_2(X) = e_+(X) - e_-(X).
\]
Recall $e_+:=\E[e_+(X)]$ and $e_-:=\E[e_-(X)]$.
Let $\text{Cov}_{\mathcal D}(A,B):=\E[(A-\E[A])(B-\E[B])]$ be the covariance between random variables $A$ and $B$ w.r.t. the distribution $\mathcal D$.
The exact effects of IDN on peer loss functions are revealed in Theorem~\ref{thm:covpeerBinary} and proved in Appendix~\ref{proof:covpeerBinary}.

\begin{theorem}[Decoupling binary IDN]
\label{thm:covpeerBinary}
In binary classifications, the expected peer loss with IDN writes as:
\begin{align}
        \E_{\widetilde{\mathcal D}}[{\BR_{\text{PL}}}(f(X),\widetilde{Y})] &=(1-e_+-e_-)\E_{\mathcal D^*}[{\BR_{\text{PL}}}(f(X),Y^*)]\notag\\
&+\text{Cov}_{\mathcal D^*}(Z_1(X),\BR(f(X),Y^*)) \notag \\
&+\text{Cov}_{\mathcal D^*}(Z_2(X),\BR(f(X),-1)). \label{eq:binaryCAP}
\end{align}
\end{theorem}

Theorem~\ref{thm:covpeerBinary} effectively divides the instance-dependent label noise into two parts.
As shown in Eq.~(\ref{eq:binaryCAP}), the first line is the same as Eq.~(\ref{eq:peerBinary}) in Lemma~\ref{lem:invariant_d}, indicating the average effect of instance-dependent label noise can be treated as a class-dependent one with parameters $e_+,e_-$. The additional two covariance terms in the second and the third lines of Eq.~(\ref{eq:binaryCAP}) characterize the additional contribution of examples due to their differences in the label noise rates. The covariance terms will become larger for a setting with more diverse noise rates, capturing a more heterogeneous and uncertain learning environment. %
Interested readers are also referred to the high-level intuitions for using covariance terms at the end of Section~\ref{sec:down-weight}. %

\rev{We now briefly discuss one extension of Theorem~\ref{thm:covpeerBinary} to a $K$-class classification task.
Following the assumption adopted in \cite{liu2019peer}, we consider a particular setting of IDN whose the expected transition matrix satisfies $T_{i,j}=T_{k,j},\, \forall i\ne j\ne k$.}
Denote by $e_j=T_{i,j}, \forall i\ne j$.
\rev{Corollary~\ref{cor:covpeerMul} decouples the effects of IDN in multi-class cases and is proved in Appendix~\ref{proof:covpeerMul}.}
\begin{corollary}[Decoupling multi-class IDN]
\label{cor:covpeerMul}
In multi-class classifications, when the expected transition matrix satisfies $e_j = T_{i,j}=T_{k,j},\, \forall i\ne j\ne k$,  the expected peer loss with IDN writes as:
\begin{align*}
&\E_{\widetilde{\mathcal D}}[\ell_{\text{PL}}(f(X),\widetilde{Y})] =(1-\sum_{i\in[K]}e_i)\E_{\mathcal D^*}[\ell_{\text{PL}}(f(X),Y^*)]\\
& + \sum_{j\in[K]}  \mathbb E_{\mathcal D_{Y^*}} \left[\text{Cov}_{\mathcal D^*|Y^*} \left(T_{Y^*,j}(X), \ell(f(X), j)\right)\right],
\end{align*}
where $\mathcal D_{Y^*}$ is the marginal distribution of $Y^*$ and $\mathcal D^*|Y^*$ is the conditional distribution of $\mathcal D^*$ given $Y^*$.  %
\end{corollary}

\subsection{Using Second-Order Statistics}

Inspired by Theorem \ref{thm:covpeerBinary}, if $\mathcal D^*$ is available, we can subtract two covariance terms and make peer loss invariant to IDN.
Specifically, define 
{\small
\begin{align*}
 \tilde f^*_{\text{\covpeer{}}} = \argmin_{f}  \E_{\widetilde{\mathcal D}}[{\BR_{\text{PL}}}(f(X),\widetilde{Y})] 
 &- \text{Cov}(Z_1(X),\BR(f(X),Y^*)) \\
 & - \text{Cov}(Z_2(X),\BR(f(X),-1)) .
\end{align*}}
We have the following optimality guarantee and its proof is deferred to Appendix~\ref{proof:optimal}. %
\begin{theorem}
\label{thm:optimal}
$\tilde f^*_{\emph{\text{\covpeer{}}}} \in \argmin_{f} ~\E_{\mathcal D^*}[\BR(f(X),Y^*)].$
\end{theorem}

For a $K$-class classification problem, a general loss function for our \textit{Covariance-Assisted Learning (\covpeer{})} approach is given by
\begin{equation}\label{eq:cov_def}
    \begin{split}
       & \ell_{\covpeer} (f(x_n),\tilde y_n) =  {\ell_{\text{PL}}} (f(x_n),\tilde y_n)\\
&- \sum_{j\in[K]}  \mathbb E_{\mathcal D_{Y^*}} \left[\text{Cov}_{\mathcal D^*|Y^*} \left(T_{Y^*,j}(X), \ell(f(X), j)\right)\right].
    \end{split}
\end{equation}
Eq.~(\ref{eq:cov_def}) shows the Bayes optimal distribution $\mathcal D^*$ is critical in implementing the proposed covariance terms.
However, $\mathcal D^*$ cannot be obtained trivially, %
and only imperfect proxy constructions of the dataset (denoted by $\estD$) could be expected.
Detailed constructions of $\estD$ are deferred to Section~\ref{sec:alg}.

\noindent\textbf{Advantages of using covariance terms}~~
There are several advantages of using the proposed covariance terms.
Unlike directly correcting labels according to $\mathcal D^*$, the proposed covariance term can be viewed as a ``soft'' correction that maintains the information encoded in both original noisy labels and the estimated Bayes optimal labels.
Keeping both information is beneficial as suggested in \cite{han2019deep}.
Moreover, compared to the direct loss correction approaches \cite{Patrini_2017_CVPR,xia2020parts,xia2019anchor}, we keep the original learning objective and apply ``correction'' using an additional term.
Our method is more robust in practice compared to these direct end-to-end loss correction approaches due to two reasons: 
1) The covariance term summarizes the impact of the complex noise using an average term, indicating that our approach is less sensitive to the estimation precision of an individual example; 2) As will be shown in Section~\ref{sec:imperfectCov}, the proposed method is tolerant with accessing an imperfect $\mathcal D^*$.

Estimating the covariance terms relies on samples drawn from distribution $\mathcal D^*$.
Thus, we need to construct a dataset $\estD$, which is similar or unbiased w.r.t. $D^*$.
We will first show the algorithm for constructing $\estD$, then provide details for DNN implementations.

\subsubsection{Constructing $\estD$}\label{sec:alg}

To achieve unbiased estimates of the variance terms, the high-level intuition for constructing $\estD$ is determining whether the label of each example in $\widetilde D$ is Bayes optimal or not by comparing the likelihood, confidence, or loss of classifying the (noisy) label to some thresholds.
There are several methods for constructing $\estD$: distillation \cite{cheng2017learningdistill}, searching to exploit \cite{yao2020searching}, and sample sieve \cite{sieve2020}.
If the model does not overfit the label noise and learns the noisy distribution, both methods in \cite{cheng2017learningdistill} and \cite{yao2020searching} work well. However, for the challenging instance-dependent label noise, overfitting occurs easily thus techniques to avoid overfitting are necessary. In this paper, we primarily adapt the sample sieve proposed in \cite{sieve2020}, which uses a confidence regularizer to avoid overfitting, to construct $\estD$. 
{Specifically, as shown in \cite{sieve2020}, in each epoch $t$, the regularized loss for each example is adjusted by the parameter $\alpha_{n,t}$, which \rev{can be calculated based on model predictions in linear time}. In the ideal cases assumed in \cite{sieve2020}, any example with a positive adjusted loss is corrupted (with a wrong label).}

\begin{algorithm}[t]
\LinesNumbered
\SetAlgoLined
\KwIn{Noisy dataset $\widetilde D$. Thresholds $L_{\min} \le L_{\max}$. Number of epochs $T$. $\estD = \widetilde D$.}
Train the sample sieve in \cite{sieve2020} for $T$ epochs and get the model $f$\;
    \For{$n\in[N]$}{
        Calculate $\alpha_{n,T}$ following \cite{sieve2020}\;
        \uIf{$\ell_{\text{CORES}^2}(f(x_n),\tilde y_n)-\alpha_{n,T} \le L_{\min}$}{
           $\hat y_n = \tilde y_n$\; \label{line:small_loss}}
           \uElseIf{$\ell_{\text{CORES}^2}(f(x_n),\tilde y_n)-\alpha_{n,T} > L_{\max}$}
           {$\hat y_n = \argmax_{y\in[K]} f_{x_n}[y]$\; \label{line:large_loss}}
           \uElse{$\hat y_n = -1$ (drop example $n$)\;\label{Line:moderate_loss}}
     } 
 \KwOut{$\estD:=\{(x_n,\hat y_n): n\in[N], \hat y_n \ne -1\}$}
 \caption{Constructing $\estD$}
 \label{alg:D}
\end{algorithm}
We summarized the corresponding procedures in Algorithm~\ref{alg:D}, where the critical thresholds for comparing losses are denoted by $L_{\min}$ and $L_{\max}$.
At Line~\ref{line:small_loss}, if the loss adjusted by $\alpha_{n,t}$ is small enough (\underline{smaller than the threshold $L_{\min}$}), we assume $\tilde y_n$ is the Bayes optimal label.
Accordingly, at Line~\ref{line:large_loss}, if the adjusted loss is too large (\underline{larger than the threshold $L_{\max}$}), we treat $\tilde y_n$ as a corrupted one and assume the class with maximum predicted probability to be Bayes optimal one.
For the examples with moderate adjusted loss, we drop it as indicated in Line~\ref{Line:moderate_loss}.
In ideal cases with infinite model capacity and sufficiently many examples (as assumed in \cite{sieve2020}), we can set thresholds $L_{\min} = L_{\max} = 0$ to guarantee a separation of clean and corrupted examples, thus $\estD$ will be an unbiased proxy to $\mathcal D^*$ \footnote{In the ideal case as assumed in Corollary 1 of \cite{sieve2020}, we have $\estD = D^*$.}.
However, in real experiments, when both the model capacity and the number of examples are limited, we may need to tune $L_{\min}$ and $L_{\max}$ to obtain a high-quality construction of $\estD$.
In this paper, we set $L_{\min} = L_{\max}$ to ensure $|\estD| = |D^*|$ and reduce the effort to tuning both thresholds simultaneously.

{Note that using $\estD$ to estimate the covariance terms could be made theoretically more rigorous by applying appropriate re-weighting techniques \cite{cheng2017learningdistill,fang2020rethinking,huang2007correcting}.
See Appendix~\ref{dis:L} for more discussions and corresponding guarantees. We omit the details here due to the space limit}. Nonetheless,
our approach is tolerant of an imperfect $\estD$, which will be shown theoretically in Section \ref{sec:imperfectCov}.

\subsubsection{Implementations}

For implementations with deep neural network solutions, we need to estimate the transition matrix $T(X)$ relying on $\estD$ and estimate the covariance terms along with stochastic gradient descent (SGD) updates.

\noindent\textbf{Covariance Estimation in SGD}~~
As required in (\ref{eq:cov_def}), with a particular $\estD$, \rev{each computation for $\hat T_{i,j}(x_n)$ requires only one time check of the associated noisy label} as follows:
\begin{equation}\label{eq:est_T}
    \hat T_{i,j}(x_n) = \BR\{ \hat y_n = i, \tilde y_n = j \}. 
\end{equation}
When $\estD$ is unbiased w.r.t. $D^*$, the estimation in (\ref{eq:est_T}) is also unbiased because
\begin{align*}
   & \E_{\widetilde{\mathcal D}|X, \hat Y=i}[\hat T_{i,j}(X)]  
  =  \E_{\widetilde{\mathcal D}|X, \hat Y=i}[\BR\{ \hat Y = i, \widetilde Y = j| X \}] \\
  = & \PP(\widetilde Y = j| X, \hat Y = i)   =  \PP(\widetilde Y = j| X, Y^* = i).
\end{align*}
Noting $\text{Cov}_{\mathcal D}(A,B):=\E[(A-\E[A])(B-\E[B])] = \E[(A-\E[A])\cdot B],$
the covariance can be estimated empirically as
\[
\frac{1}{N}\sum_{n\in[N]}\sum_{i,j \in [K]}   \BR\{y_n^*=i\} \left[(\hat T_{i,j}(x_n) - \hat T_{i,j}) \cdot \ell (f(x_n), j) \right].
\]
\rev{For each batch of data, the above estimation has $O(N)$ complexities in computation and space. To reduce
both complexities, with the cost of the estimation quality, we use $|E_b|$ examples to estimate the covariance in each batch, where $E_b$ is the set of sample indices of batch-$b$. Per sample wise, Eq.~(\ref{eq:cov_def}) can be transformed to}
\begin{equation*}%
    \begin{split}
        &\ell_{\covpeer} (f(x_n),\tilde y_n) =  {\ell_{\text{PL}}} (f(x_n),\tilde y_n) \\
        & -  \sum_{i,j\in[K]}  \BR\{y_n^*=i\} \left[(\hat T_{i,j}(x_n) - \hat T_{i,j}) \cdot \ell (f(x_n), j) \right].
    \end{split}
\end{equation*}
\rev{With the above implementation, the estimation is done locally for each point in $O(1)$ complexity.}
\subsection{\covpeer{} with Imperfect Covariance Estimates} 
\label{sec:imperfectCov}

As mentioned earlier, $\mathcal{D}^*$ cannot be perfectly obtained in practice. Thus, there is a performance gap between the ideal case (with perfect knowledge of $\mathcal D^*$) and the actually achieved one.
We now analyze the effect of imperfect covariance terms (Theorem~\ref{thm:imperfect}).

Denote the imperfect covariance estimates by $\estDg$, where $\tau\in[0,1]$ is the expected ratio (a.k.a. probability) of correct examples in $\estDg$:
$\tau = \E[\BR \{ (X,\hat Y) \in \estDg | (X,Y^*) \in D^* \}] =  \PP( (X,\hat Y)\sim \estDisg | (X,Y^*)\sim \mathcal D^* ).$
With $\estDg$, the minimizer of the 0-1 \covpeer{} loss is given by: 
{\small
\begin{align*}
    \tilde f^*_{\text{\covpeer{}-}\tau} \hspace{-2pt} = \hspace{-2pt}\argmin_{f}
& ~ \E_{\widetilde{\mathcal D}} \bigg[{\BR_{\text{PL}}}(f(X),\widetilde{Y})] \hspace{-2pt} - \hspace{-2pt} \text{Cov}_{\estDisg}(Z_1(X),\BR(f(X),\hat Y)) \\
& - \text{Cov}_{\estDisg}(Z_2(X),\BR(f(X),-1))\bigg].
\end{align*}}
Theorem~\ref{thm:imperfect} reports the error bound produced by $\tilde f^*_{\text{\covpeer{}-}\tau}$. See Appendix~\ref{proof:imperfect} for the proof.
\begin{theorem}[Imperfect Covariance]\label{thm:imperfect}
With $\estDg$, when $p^*=0.5$, we have
\[
\E[\BR(\tilde f^*_{\text{\covpeer{}-}\tau}(X),Y^*)] \leq \frac{4(1-\tau)(\epsilon_+ + \epsilon_-)}{1-e_+-e_-}.
\]
\end{theorem}

Theorem~\ref{thm:imperfect} shows the quality of $\estDg$ controls the scale of the worst-case error upper-bound.
Compared with Theorem~\ref{thm:peerIDN} where no covariance term is used, we know the covariance terms will always be helpful when $\tau \in [0.5,1]$. That is, the training with the assistance of covariance terms will achieve better (worst-case) accuracy on the Bayes optimal distribution when the construction $\estDg$ is better than a dataset that includes each instance in $D^*$ randomly with 50\% chance.

%% file: src/exp.tex
\section{Experiments}
We now present our experiment setups and results. 
\subsection{General Experiment Settings}

\noindent\textbf{Datasets and models}~~
The advantage of introducing our second-order approach is evaluated on three benchmark datasets: CIFAR10, CIFAR100 \cite{krizhevsky2009learning} and Clothing1M \cite{xiao2015learning}. Following the convention from \cite{sieve2020,xu2019l_dmi}, we use ResNet34 for CIFAR10 and CIFAR100 and ResNet50 for Clothing1M.
Noting the expected peer term  $\E_{\mathcal{D}_{\widetilde{Y}|\widetilde{D}}}[\ell(f(x_n),\widetilde Y)]$ (a.k.a. \textit{confidence regularizer (CR)} as implemented in \cite{sieve2020}) is more stable and converges faster than the one with peer samples, we train with $\ell_{\text{CORES}^2}$.
It also enables a fair ablation study since $\estD$ is constructed relying on \cite{sieve2020}.
For numerical stability, we use a cut-off version of the cross-entropy loss $\ell(f(x),y) = -\ln(f_x[y]+\varepsilon)$.
Specifically, we use $\varepsilon = 10^{-8}$ for the traditional cross-entropy term, use $\varepsilon = 10^{-5}$ for the CR term, and the covariance term.
All the experiments use a momentum of $0.9$.
The weight decay is set as $0.0005$ for CIFAR experiments and $0.001$ for Clothing1M.

\noindent\textbf{Noise type}~~
For CIFAR datasets, the instance-dependent label noise is generated following the method from \cite{sieve2020,xia2020parts}.
\rev{The basic idea is randomly generating one vector for each class ($K$ vectors in total) and project each incoming  feature  onto  these $K$ vectors.
The label noise is added by jointly considering the clean label and the projection  results.
See Appendix \ref{sec:instance_noise_gen} for details.}
In expectation, the noise rate $\eta$ is the overall ratio of examples with a wrong label in the entire dataset.
For the Clothing1M dataset, we train on 1 million noisy training examples that encode the real-world human noise. 

\subsection{Baselines}
We compare our method with several related works, where the cross-entropy loss is tested as a common baseline.
Additionally, the generalized cross-entropy \cite{zhang2018generalized} is compared as a generalization of mean absolute error and cross-entropy designed for label noise.
Popular loss correction based methods \cite{patrini2017making,xia2020parts,xia2019anchor}, sample selection based methods \cite{sieve2020,han2018co,wei2020combating,yu2019does}, and noise-robust loss functions \cite{liu2019peer,xu2019l_dmi} are also chosen for comparisons.
All the compared methods adopt similar data augmentations, including standard random crop, random flip, and normalization.
Note the recent work on part-dependent label noise \cite{xia2020parts} did not apply random crop and flip on the CIFAR dataset.
For a fair comparison with \cite{xia2020parts}, we remove the corresponding data augmentations from our approach and defer the comparison to Appendix~\ref{sec:noaug}.
The semi-supervised learning based methods with extra feature-extraction and data augmentations are not included.
\rev{All the CIFAR experiments are repeated $5$ times with independently synthesized IDN.}
The highest accuracies %
on the clean testing dataset are \rev{averaged over $5$ trials to show the best generalization ability of each method.}

\subsection{Performance Comparisons}
\subsubsection{CIFAR}

\begin{table*}[!t]
		\caption{Comparison of test accuracies ($\%$) using different methods.}
		\begin{center}
		\scalebox{.8}{{\begin{tabular}{c|cccccc} 
				\hline 
				 \multirow{2}{*}{Method}  & \multicolumn{3}{c}{\emph{Inst. CIFAR10} } & \multicolumn{3}{c}{\emph{Inst. CIFAR100} } \\ 
				 & $\eta = 0.2$&$\eta = 0.4$&$\eta = 0.6$ & $\eta = 0.2$&$\eta = 0.4$&$\eta = 0.6$\\
				\hline\hline
			     CE (Standard)  &85.45$\pm$0.57 & 76.23$\pm$1.54 &  59.75$\pm$1.30 &  57.79$\pm$1.25 & 41.15$\pm$0.83 & 25.68$\pm$1.55 \\
				 Forward $T$ \cite{patrini2017making} & 87.22$\pm$1.60 & 79.37$\pm$2.72 & 66.56$\pm$4.90  & 58.19$\pm$1.37 & 42.80$\pm$1.01 & 27.91$\pm$3.35\\
				 $L_{\sf DMI}$ \cite{xu2019l_dmi}  &88.57$\pm$0.60 & 82.82$\pm$1.49 & 69.94$\pm$1.31 & 57.90$\pm$1.21 & 42.70$\pm$0.92 & 26.96$\pm$2.08\\
				 $L_{q}$ \cite{zhang2018generalized}   & 85.81$\pm$0.83 & 74.66$\pm$1.12 & 60.76$\pm$3.08  & 57.03$\pm$0.27 & 39.81$\pm$1.18 & 24.87$\pm$2.46\\
				 Co-teaching \cite{han2018co} & 88.87$\pm$0.24 & 73.00$\pm$1.24 & 62.51$\pm$1.98 & 43.30$\pm$0.39 & 23.21$\pm$0.57 & 12.58$\pm$0.51\\
				 Co-teaching+ \cite{yu2019does}  & 89.80$\pm$0.28 & 73.78$\pm$1.39 & 59.22$\pm$6.34 & 41.71$\pm$0.78 & 24.45$\pm$0.71 & 12.58$\pm$0.51\\
				JoCoR \cite{wei2020combating} & 88.78$\pm$0.15 & 71.64$\pm$3.09 & 63.46$\pm$1.58 & 43.66$\pm$1.32 & 23.95$\pm$0.44 & 13.16$\pm$0.91\\
				Reweight-R \cite{xia2019anchor} & 90.04$\pm$0.46 & 84.11$\pm$2.47 & 72.18$\pm$2.47 & 58.00$\pm$0.36 & 43.83$\pm$8.42 & 36.07$\pm$9.73\\
				Peer Loss \cite{liu2019peer} &89.12$\pm$0.76 & 83.26$\pm$0.42 & 74.53$\pm$1.22  & 61.16$\pm$0.64 & 47.23$\pm$1.23 & 31.71$\pm$2.06\\
				\SPL{} \cite{sieve2020} & 91.14$\pm$0.46 & 83.67$\pm$1.29 & 77.68$\pm$2.24 &66.47$\pm$0.45 & 58.99$\pm$1.49 & 38.55$\pm$3.25\\
				\covpeer{}  & \textbf{92.01$\pm$0.75} & \textbf{84.96$\pm$1.25} & \textbf{79.82$\pm$2.56}  & \textbf{69.11$\pm$0.46} & \textbf{63.17$\pm$1.40} & \textbf{43.58$\pm$3.30}\\
			   \hline
			\end{tabular}}}
		\end{center}
		\vspace{-8pt}
		\label{table:cifar-inst}
\end{table*}

In experiments on CIFAR datasets, we use a batch size of $128$, an initial learning rate of $0.1$, and reduce it by a factor of $10$ at epoch $60$.

\noindent\textbf{Construct $\estD$}~~
To construct $\estD$, we update the DNN for $65$ epochs by minimizing $\ell_{\text{CORES}^2}$ (without dynamic sample sieve) and apply Algorithm~\ref{alg:D} with $L_{\min} = L_{\max} = -8$.\footnote{Theoretically, we have $L_{\min} = L_{\max} = 0$ if both the CE term and the CR term use a log loss without cut-off ($\varepsilon=0$). Current setting works well (not the best) for CIFAR experiments empirically.}
For a numerically stable solution, we use the square root of the noise prior for the CR term in $\ell_{\text{CORES}^2}$ as $- \beta   \sum_{i\in[K]} \frac{\sqrt{\PP(\widetilde Y = i|\widetilde{D})}}
{\sum_{j=1}^K \sqrt{\PP(\widetilde Y = j|\widetilde{D})}} \ell(f(x_n),i).$
The hyperparameter $\beta$ is set to $2$ for CIFAR10 and $10$ for CIFAR100.

\noindent\textbf{Train with \covpeer{}}~~
With an estimate of $D^*$, we re-train the model $100$ epochs.
The hyper-parameter $\beta$ is set to $1$ for CIFAR10 and $10$ for CIFAR100.
Note the hyperparameters ($L_{\min}$, $L_{\max}$, $\beta$) can be better set if a clean validation set is available.

\noindent\textbf{Performance}~~
Table~\ref{table:cifar-inst} compares the means and standard deviations of test accuracies on the clean test dataset when the model is trained with synthesized instance-dependent label noise in different levels. 
All the compared methods use ResNet34 as the backbone.
On CIFAR10, with a low-level label noise ($\eta=0.2$), all the compared methods perform well and achieve higher average test accuracies than the standard CE loss.
When the overall noise rates increase to high, most of the methods suffer from severe performance degradation while \covpeer{} still achieves the best performance.
There are similar observations on CIFAR100.
By comparing \covpeer{} with \SPL{}, we conclude that the adopted second-order statistics do work well and bring non-trivial performance improvement. 
Besides, on the CIFAR100 dataset with $\eta=0.4$ and $0.6$, we observe Reweight-R \cite{xia2019anchor} has a large standard deviation and a relatively high mean, indicating it may perform as well as or even better than \covpeer{} in some trials. 
It also shows the potential of using a revised transition matrix $T$ \cite{xia2019anchor} in severe and challenging instance-dependent label noise settings.

\subsubsection{Clothing1M}
For Clothing1M, we first train the model following the settings in \cite{sieve2020} and construct $\estD$ with the best model.
Noting the overall accuracy of noisy labels in Clothing1M is about $61.54\%$ \cite{xiao2015learning}, we set an appropriate $L_{\min} = L_{\max}$ such that $61.54\%$ of training examples satisfying $\ell_{\text{CORES}^2}-\alpha_{n,t} \le L_{\min}$.
With $\estD$, we sample a class-balanced dataset by randomly choosing $18,976$ noisy examples for each class and continue training the model with $\beta = 1$ and an initial learning rate of $10^{-5}$ for $120$ epochs.
Other parameters are set following \cite{sieve2020}.
See Appendix~\ref{sec:detailImplement} for more detailed experimental settings.
Table~\ref{table:c1m} shows \covpeer{} performs well in the real-world human noise.

\subsection{Ablation Study}

Table~\ref{table:analysis_component} shows either the covariance term or the peer term can work well individually and significantly improve the performance when they work jointly.
Comparing the first row with the second row, we find the second-order statistics can work well (except for $\eta=0.4$) even without the peer (CR) term. 
In row 4, we show the performance at epoch $65$ since the second-order statistics are estimated relying on the model prediction at this epoch.
By comparing row 4 with row 5, we know the second-order statistics indeed lead to non-trivial improvement in the performance.
Even though the covariance term individually can only achieve an accuracy of $78.49$ when $\eta=0.4$, it can still contribute more than $1\%$ of the performance improvement (from $84.41\%$ to $85.55\%$) when it is implemented with the peer term.
This observation shows the robustness of \covpeer{}.

\begin{table}[!t]
	\caption{The best epoch (clean) test accuracies on Clothing1M. }
	\vspace{-3pt}
	\begin{center}
	\scalebox{.8}{{
		\begin{tabular}{c|c} 
			\hline 
			Method & Accuracy \\
			\hline \hline 
			CE (standard) &  68.94\\
			Forward $T$  \cite{patrini2017making} & 70.83\\
			Co-teaching \cite{han2018co}  &69.21 \\
			JoCoR \cite{wei2020combating} &70.30 \\
			$L_{\sf DMI}$ \cite{xu2019l_dmi} & 72.46\\
			PTD-R-V\cite{xia2020parts} & 71.67 \\
			\SPL{} \cite{sieve2020} & 73.24\\
			\covpeer{}  & \textbf{74.17} \\
			\hline 
		\end{tabular}
		}}
	\end{center}
	\vspace{-5pt}
	\label{table:c1m}
\end{table}
\begin{table}[!t]
	\caption{Analysis of each component of \covpeer{} on CIFAR10. The result of a particular trial is presented. \textit{Cov.}: the covariance term. \textit{Peer}: the CR term \cite{sieve2020} (a.k.a. expected peer term \cite{liu2019peer}).}
	\vspace{-5pt}
	\begin{center}	
	\scalebox{.8}{
	\begin{tabular}{cccc|ccc} 
			\hline 
			 row \# & \emph{Cov.}  & \emph{Peer} & Epoch  & $\eta = 0.2$&$\eta = 0.4$&$\eta = 0.6$\\
			\hline\hline
			1 & \xmark&\xmark & Best & 90.47 & 82.56 & 64.65 \\
			2 & \cmark&\xmark & Best & 92.10  & 78.49 & 73.55\\
			3 & \xmark&\cmark & Best &  91.85 & 84.41 &  78.74\\
			4 & \xmark&\cmark & Fixed@65 & 90.73 & 82.76 &  77.70\\
			5 & \cmark &\cmark & Best  &92.69 & 85.55& 81.54\\
			\hline
		\end{tabular}}
	\end{center}
	\vspace{-8pt}
	\label{table:analysis_component}
\end{table}

%% file: src/conclusion.tex
\section{Conclusions}
\vspace{-2pt}
This paper has proposed a second-order approach to transforming the challenging instance-dependent label noise into a class-dependent one such that existing methods targeting the class-dependent label noise could be implemented.
Currently, the necessary information for the covariance term is estimated based on a sample selection method.
Future directions of this work include extensions to other methods for estimating the covariance terms accurately.
We are also interested in exploring the combination of second-order information with other robust learning techniques.

\noindent\textbf{Acknowledgements}~~
This research is supported in part by National Science Foundation (NSF) under grant IIS-2007951, %
and in part by Australian Research Council Projects, i.e., DE-190101473.

%% file: src/appendix.tex
\clearpage
\newpage
\appendix
\onecolumn
\noindent{\Large \bf Appendix}

\section{Proof for Lemmas}

\subsection{Proof for Lemma~\ref{lem:invariant_d}}\label{proof:invariant_d}

\begin{proof}

We try to build the connection between noisy distribution $\widetilde{\mathcal{D}}$ and the underlying Bayes optimal distribution $\mathcal D^*$ by the noise rates $e_+$ and $e_-$. 
The primary difference from the proof of Lemma 2 in \cite{liu2019peer} is the usage of $\mathcal D^*$.
Note:
\begin{align*}
  & \mathbb{E}_\mathcal{\widetilde{D}}[\ell(f(X), \widetilde{Y})] \\
= & \mathbb{E}_\mathcal{{D^*}}\left[ \sum_{j \in \{-1,+1\}} \PP(\widetilde Y = j|X,Y^*) \ell(f(X),j) \right] \\ 
= & \mathbb{E}_\mathcal{{D^*}}\left[ \sum_{j \in \{-1,+1\}} \PP(\widetilde Y = j|Y^*) \ell(f(X),j) \right] \\ 
= &  \sum_{i\in \{-1,+1\}} \PP(Y^*=i) \mathbb{E}_{\mathcal{D^*}|Y^*=i}[  \PP(\widetilde Y = +1|Y^*=i) \ell(f(X),+1)  + \PP(\widetilde Y = -1|Y^*=i) \ell(f(X),-1)]\\
= & \PP(Y^*=+1) \mathbb{E}_{\mathcal{D^*}|Y^*=+1}[  (1-e_{+}) \ell(f(X),+1)  + e_+ \ell(f(X),-1)]\\
 & + \PP(Y^*=-1) \mathbb{E}_{\mathcal{D^*}|Y^*=-1}[  (1-e_{-}) \ell(f(X),-1)  + e_- \ell(f(X),+1)].
\end{align*}
Similarly, following the proof of Lemma 2 in \cite{liu2019peer}, we can prove this lemma.
\end{proof}

\subsection{Proof for Lemma~\ref{lem:peerBayes}}\label{proof:peerBayes}

\paragraph{Peer Loss on the Bayes Optimal Distribution}

Recall our goal is to learn a classifier $f$ from the noisy distribution $\widetilde{\mathcal D}$ which also minimizes the loss on the corresponding Bayes optimal distribution $D^*$, i.e.
$\E [\BR(f(X), Y^*)], (X,Y^*)\sim \mathcal D^*$. 
Before considering the case with label noise, we need to prove peer loss functions induce the Bayes optimal classifier when minimizing the 0-1 loss on $\mathcal D^*$ as in Lemma~\ref{lem:peerBayes}.

\begin{lemma}\label{lem:peerBayes}
Given the Bayes optimal distribution $\mathcal D^*$, the optimal peer classifier defined below:
\[
f^*_{\text{peer}} = \argmin_{f} ~\E_{\mathcal D^*}[{{\BR_{\text{PL}}}}(f(X),Y^*)]
\]
also minimizes $\E_{\mathcal D^*}[\BR(f(X),Y^*)]$.
\end{lemma}
See the proof below.
It has been shown in \cite{liu2019peer} that Lemma~\ref{lem:peerBayes} holds for the clean distribution $\mathcal D$ when the clean dataset is class-balanced, i.e. $\PP(Y=-1) = \PP(Y=+1) = 0.5$. For the Bayes optimal distribution $\mathcal D^*$, as shown in Lemma~\ref{lem:peerBayes}, there is \emph{no requirement} for the prior $p^*:=\PP(Y^*=+1)$.

\begin{proof}

Recall $Y^*$ is the Bayes optimal label defined as 
\[
Y^*|X := \argmax_Y ~\PP(Y|X), (X,Y)\sim \mathcal D.
\]
We need to prove that the ``optimal peer classifier" defined below:
\[
f^*_{\text{peer}} = \argmin_{f} ~\E_{\mathcal D^*}[{\BR_{\text{PL}}}(f(X),Y^*)]
\]
is the same as the Bayes optimal classifier $f^*$. To see this, suppose the claim is wrong. Denote by (notations $\epsilon_+$ and $\epsilon_-$ are defined only for this proof): 
\[
\epsilon_+ := \PP(f^*_{\text{peer}}(X) =-1|f^*(X)=+1), ~~
\epsilon_- := \PP(f^*_{\text{peer}}(X) =+1|f^*(X)=-1)
\]
and denote by $p^* := \PP(f^*(X)=+1)$.
Then 
\begin{align*}
    &\E_{\mathcal D^*}[{\BR_{\text{PL}}}(f^*_{\text{peer}}(X),Y^*)]\\
    &= \PP(f^*_{\text{peer}}(X) \neq Y^*) - p^* \cdot \PP(f^*_{\text{peer}}(X) \neq +1) - (1-p^*) \cdot \PP(f^*_{\text{peer}}(X) \neq -1)\\
    &= p^* \cdot \epsilon_+ + (1-p^*) \cdot \epsilon_- - p^* \cdot \PP(f^*_{\text{peer}}(X) \neq +1) - (1-p^*) \cdot \PP(f^*_{\text{peer}}(X) \neq -1)\\
    &=p^* \cdot \epsilon_+ + (1-p^*) \cdot \epsilon_-  \\
    &- p^* \cdot \left( \PP(f^*_{\text{peer}}(X) \neq +1|f^*(X) \neq +1) \PP(f^*(X) \neq +1)+ \PP(f^*_{\text{peer}}(X) \neq +1|f^*(X) \neq -1) \PP(f^*(X) \neq -1) \right)\\
    &- (1-p^*) \cdot \left( \PP(f^*_{\text{peer}}(X) \neq -1|f^*(X) \neq +1) \PP(f^*(X) \neq +1)+ \PP(f^*_{\text{peer}}(X) \neq -1|f^*(X) \neq -1) \PP(f^*(X) \neq -1)\right)\\
    &=p^* \cdot \epsilon_+ + (1-p^*) \cdot \epsilon_-  \\
    &- p^* \cdot \PP(f^*(X) \neq +1) (1-\epsilon_-) -p^* \cdot \PP(f^*(X) \neq -1) \cdot \epsilon_+\\
    &-(1-p^*) \cdot \PP(f^*(X) \neq -1) (1-\epsilon_+) - (1-p^*) \cdot \PP(f^*(X) \neq +1) \cdot \epsilon_-\\
    &=0 -  p^* \cdot \PP(f^*(X) \neq +1) -(1-p^*) \cdot \PP(f^*(X) \neq -1) \\
    &+p^*(\epsilon_+  + \PP(f^*(X) \neq +1) \epsilon_- -\PP(f^*(X) \neq -1) \epsilon_+ )\\
    &+(1-p^*) (\epsilon_-  + \PP(f^*(X) \neq -1) \epsilon_+ -\PP(f^*(X) \neq +1) \epsilon_- )\\
    &> 0 -  p^* \cdot \PP(f^*(X) \neq +1) -(1-p^*) \cdot \PP(f^*(X) \neq -1) \\
    &=\E_{\mathcal D^*}[{\BR_{\text{PL}}}(f^*(X),Y^*)]
\end{align*}
contradicting the optimality of $f^*_{\text{peer}}$. Thus our claim is proved.
\end{proof}
\section{Proof for Theorems}

\subsection{Proof for Theorem~\ref{thm:peerIDN}}\label{proof:peerIDN}

\begin{proof}
The covariance $\text{Cov}(\cdot,\cdot)$ in this proof is taken over the Bayes optimal distribution $\mathcal D^*$.
The following proof is built on the result of Theorem~\ref{thm:covpeerBinary}, i.e. Eq.~(\ref{eq:binaryCAP}).
First note
\begin{align*}
     \text{Cov}(Z_1(X),\BR(f_1(X),Y^*) - \BR(f_2(X),Y^*)) & = \E[(Z_1(X)-\E[Z_1(X)])\cdot (\BR(f_1(X),Y^*)- \BR(f_2(X),Y^*))]\\
    &\leq \E[|(Z_1(X)-\E[Z_1(X)]|] \\
    &\leq \E|e_+(X)-\E[e_+(X)]|+\E|e_-(X)-\E[e_-(X)]|
\end{align*}
Similarly, one can show that 
\begin{align*}
        \text{Cov}(Z_2(X),\BR(f_1(X),-1) - \BR(f_2(X),-1)) \leq \E|e_+(X)-\E[e_+(X)]|+\E|e_-(X)-\E[e_-(X)]|
\end{align*}

Now with bounded variance in the error rates, suppose: 
\[
\E|e_+(X)-\E[e_+(X)]|\leq \epsilon_+, \quad \E|e_-(X)-\E[e_-(X)]| \leq \epsilon_-
\]
Note
\begin{equation*}
\begin{split}
    \tilde f^*_{\text{peer}}  := &\argmin_{f} \E_{\widetilde {\mathcal D}} \left[ {\BR_{\text{PL}}}(f(X), \tilde Y)  \right]\\
     = & \argmin_{f} \left[(1-e_+-e_-) \E_{\mathcal D^*}[{\BR_{\text{PL}}}(f(X),Y^*) + \text{Cov}(Z_1(X),\BR(f(X),Y^*)) + \text{Cov}(Z_2(X),\BR(f(X),-1)) \right]\\
     = & \argmin_{f} \big[(1-e_+-e_-) \left( \E_{\mathcal D^*}[\BR(f(X),Y^*) - p^* \cdot\E_{\mathcal D^*}[\BR(f(X),+1)] - (1 - p^* )\cdot\E_{\mathcal D^*}[\BR(f(X),-1)] \right)\\
      & \qquad \qquad  + \text{Cov}(Z_1(X),\BR(f(X),Y^*)) + \text{Cov}(Z_2(X),\BR(f(X),-1)) \big].
\end{split}
\end{equation*}
Then
\begin{equation*}
    \begin{split}
        & \E_{\mathcal D^*} \left[ \BR(\tilde f^*_{\text{peer}}(X),Y^*)  \right] + \frac{\text{Cov}(Z_1(X),\BR(\tilde f^*_{\text{peer}}(X),Y^*)) + \text{Cov}(Z_2(X),\BR(\tilde f^*_{\text{peer}}(X),-1)) }{1-e_+-e_-}   \\
     =  & \E_{\mathcal D^*} \left[ \BR(\tilde f^*_{\text{peer}}(X),Y^*)  \right]- 0.5 \cdot \E_X[ \BR(\tilde f^*_{\text{peer}}(X),+1) ] - 0.5 \cdot \E_X[ \BR(\tilde f^*_{\text{peer}}(X),-1) ] + 0.5 \\
     & + \frac{\text{Cov}(Z_1(X),\BR(\tilde f^*_{\text{peer}}(X),Y^*)) + \text{Cov}(Z_2(X),\BR(\tilde f^*_{\text{peer}}(X),-1)) }{1-e_+-e_-} \\
    \le & \E_{\mathcal D^*} \left[ \BR(\tilde f^*_{\text{peer}}(X),Y^*)  \right]  - p^* \cdot \E_X[ \BR(\tilde f^*_{\text{peer}}(X),+1) ] - (1-p^*) \cdot \E_X[ \BR(\tilde f^*_{\text{peer}}(X),-1) ] + |p^*-0.5| + 0.5 \\
    & +  \frac{\text{Cov}(Z_1(X),\BR(\tilde f^*_{\text{peer}}(X),Y^*)) + \text{Cov}(Z_2(X),\BR(\tilde f^*_{\text{peer}}(X),-1)) }{1-e_+-e_-}\\
    \le &  \E_{\mathcal D^*} \left[ \BR(f^*(X),Y^*)  \right] - p^* \cdot \E_X[ \BR(f^*(X),+1) ] - (1-p^*) \cdot \E_X[ \BR(f^*(X),-1) ] + |p^*-0.5| + 0.5 \\
    & + \frac{\text{Cov}(Z_1(X),\BR(f^*(X),Y^*)) + \text{Cov}(Z_2(X),\BR(f^*(X),-1))}{1-e_+-e_-} \\
    \le &  \E_{\mathcal D^*} \left[ \BR(f^*(X),Y^*)  \right]+ \frac{\text{Cov}(Z_1(X),\BR(f^*(X),Y^*)) + \text{Cov}(Z_2(X),\BR(f^*(X),-1))}{1-e_+-e_-} + 2|p^*-0.5|.
    \end{split}
\end{equation*}%
Thus 
\begin{equation*}
    \begin{split}
        & \E_{\mathcal D^*} \left[ \BR(\tilde f^*_{\text{peer}}(X),Y^*) - \BR(f^*(X),Y^*)  \right] \\
    \le & \frac{\text{Cov}(Z_1(X),\BR(f^*(X),Y^*)-\BR(\tilde f^*_{\text{peer}}(X),Y^*)) + \text{Cov}(Z_2(X),\BR(f^*(X),-1)-\BR(\tilde f^*_{\text{peer}}(X),-1))}{1-e_+-e_-} + 2|p^*-0.5| \\
    \le & 2 \frac{\E|e_+(X)-\E[e_+(X)]|+\E|e_-(X)-\E[e_-(X)]| }{1-e_+-e_-} + 2|p^*-0.5| \\
    \le & \frac{2(\epsilon_+ + \epsilon_-)}{1-e_+-e_-} + 2|p^*-0.5|.
    \end{split}
\end{equation*}
Noting $\BR(f^*(X),Y^*) =0$, we finish the proof.

\end{proof}

\subsection{Proof for Theorem~\ref{thm:covpeerBinary}}\label{proof:covpeerBinary}
\begin{proof}

The covariance $\text{Cov}(\cdot,\cdot)$ in this proof is taken over the Bayes optimal distribution $\mathcal D^*$.
Recall 
\[
e_+(X) := \PP(\widetilde{Y}=-1|Y^* = +1, X),
e_-(X) := \PP(\widetilde{Y}=+1|Y^* = -1, X)
\]
and 
\[
e_+ := \E_X [e_+(X)],~~ e_- := \E_X[e_-(X)]
\]
We first have the following equality:
\begin{align*}
\E_{\widetilde{\mathcal D}}[{\BR_{\text{PL}}}(f(X),\tilde{Y})] &= \E_{\mathcal D^*}[(1-e_+(X)-e_-(X)) \BR(f(X),Y^*))] \qquad \qquad \text{(Term-A)}\\
&+ \E_{X}[e_+(X)\BR(f(X),-1) + e_-(X)\BR(f(X),+1) ]   \qquad \text{(Term-B)}\\
&-(1-e_+-e_-) \cdot \E_{D^*}[\BR(f(X),Y^*_p))] \qquad\qquad\qquad~~~ \text{(Term-C)}\\
&-\E_{X}[e_+ \cdot \BR(f(X),-1) + e_- \cdot \BR(f(X),+1) ]  \qquad \qquad \text{(Term-D)}
\end{align*}
Term-B can be transformed to:
\begin{align*}
     &\E_{X}[e_+(X) \cdot \BR(f(X),-1) + e_-(X)\cdot \BR(f(X),+1) ]\\
     &= \E_{X}[e_+(X) \cdot \BR(f(X),-1) + e_-(X)\cdot (1-\BR(f(X),-1)) ]\\
     &=\E_{X}[(e_+(X)-e_-(X)) \cdot \BR(f(X),-1) + e_-(X)].
\end{align*}
Similarly, Term-D turns to
\[
\E_{X}[e_+ \cdot \BR(f(X),-1) + e_- \cdot \BR(f(X),+1) ] = (e_+ - e_-) \cdot \E_{X}[\BR(f(X),-1)] + e_-.
\]
Define two random variables
\[
Z_1(X):= 1-e_+(X)-e_-(X), ~Z_2(X) = e_+(X) - e_-(X).
\]
Then Term-A becomes
\begin{align*}
    &\E_{\mathcal D^*}[(1-e_+(X)-e_-(X)) \BR(f(X),Y^*))] \\
    &= \E[Z_1(X)] \cdot   \E_{\mathcal D^*}[ \BR(f(X),Y^*))] + \text{Cov}(Z_1(X),\BR(f(X),Y^*))\\
    &=(1-e_+-e_-) \cdot \E_{\mathcal D^*}[ \BR(f(X),Y^*))] + \text{Cov}(Z_1(X),\BR(f(X),Y^*))
\end{align*}
Similarly, Term-B can be further transformed to
\begin{align*}
    &\E_{X}[(e_+(X)-e_-(X)) \cdot \BR(f(X),-1) + e_-(X)] \\
    &= \E[Z_2(X)] \E_X[\BR(f(X),-1)] + \text{Cov}(Z_2(X),\BR(f(X),-1)) + e_-\\
    &= (e_+ - e_-) \E_X[\BR(f(X),-1)] + \text{Cov}(Z_2(X),\BR(f(X),-1)) + e_-
\end{align*}
Combining the above results, we have
\begin{align*}
\E_{\widetilde{\mathcal D}}[{\BR_{\text{PL}}}(f(X),\tilde{Y})] &= (1-e_+-e_-) \cdot  \E_{\mathcal D^*}[\BR(f(X),Y^*))] \\
&+ (e_+ - e_-) \E_X[\BR(f(X),-1)] + e_-\\
&-(1-e_+-e_-) \cdot \E_{\mathcal D^*}[\BR(f(X),Y^*_p))] \\
&-(e_+ - e_-) \cdot \E_{X}[\BR(f(X),-1)] - e_- \\
&+\text{Cov}(Z_1(X),\BR(f(X),Y))+ \text{Cov}(Z_2(X),\BR(f(X),-1)) \\
&=(1-e_+-e_-)\E_{\mathcal D^*}[{\BR_{\text{PL}}}(f(X),Y^*)]\\
&+\text{Cov}(Z_1(X),\BR(f(X),Y^*))+ \text{Cov}(Z_2(X),\BR(f(X),-1)) \\
\end{align*}
\end{proof}

\subsection{Proof for Theorem~\ref{thm:optimal}}\label{proof:optimal}

\begin{proof}
From Theorem~\ref{thm:covpeerBinary}, we know
\begin{align*}
&\mathbb E_{\widetilde{\mathcal {D}}}[{\BR_{\text{PL}}}(f(X), \widetilde{Y})] - \text{Cov}(Z_1(X),\BR(f(X), Y^*))  - \text{Cov}(Z_2(X),\BR(f(X),-1))\bigg] \\
=& (1-e_{-}-e_{+}) \cdot \mathbb E_{\mathcal D^*}[{\BR_{\text{PL}}}(f(X), Y^*)].
   \end{align*}
With Lemma~\ref{lem:peerBayes}, we can finish the proof.
\end{proof}

\subsection{Proof for Theorem~\ref{thm:imperfect}}\label{proof:imperfect}
\begin{proof}

Recall $\tau\in[0,1]$ is the expected ratio (a.k.a. probability) of correct examples in $\estDg$, i.e.
$\tau = \E[\BR \{ (X,\hat Y) \in \estDg | (X,Y^*) \in D^* \}] =  \PP( (X,\hat Y)\sim \estDisg | (X,Y^*)\sim \mathcal D^* ).$
With $\estDg$, the classifier learned by minimizing the 0-1 \covpeer{} loss is 
{
\begin{align*}
    \tilde f^*_{\text{\covpeer{}-}\tau}  := \argmin_{f}
& ~ \E_{\widetilde{\mathcal D}} \bigg[{\BR_{\text{PL}}}(f(X),\widetilde{Y})]  - \text{Cov}_{\estDisg}(Z_1(X),\BR(f(X),\hat Y))  - \text{Cov}_{\estDisg}(Z_2(X),\BR(f(X),-1))\bigg].
\end{align*}}
Note
\begin{align*}
    \text{Cov}_{\estDisg}(Z_1(X),\BR(f(X),Y)) 
    &= \E_{\estDisg} \left[
    \left( Z_1(X) - \E_{\estDisg}[Z_1(X)]  \right)\left( \BR(f(X),Y) - \E_{\estDisg} [\BR(f(X),Y)]  \right)\right] \\
    &= \E_{\estDisg} \left[
    \left( Z_1(X) - \E_{\estDisg}[Z_1(X)]  \right) \BR(f(X),Y)\right] \\
    &= \PP((X,Y)\in D^*|(X,Y) \in \estDg) \E_{\estDisg} \left[
    \left( Z_1(X) - \E_{\estDisg}[Z_1(X)]  \right) \BR(f(X),Y)|(X,Y)\in D^*\right] \\
    &+\PP((X,Y)\notin D^*|(X,Y) \in \estDg) \E_{\estDisg} \left[
    \left( Z_1(X) - \E_{\estDisg}[Z_1(X)]  \right) \BR(f(X),Y)|(X,Y)\notin D^*\right].
\end{align*}
Similarly,
\begin{align*}
    \text{Cov}_{\mathcal D^*}(Z_1(X),\BR(f(X),Y)) 
    &= \E_{\mathcal D^*} \left[
    \left( Z_1(X) - \E_{\mathcal D^*}[Z_1(X)]  \right) \BR(f(X),Y)\right] \\
    &= \PP((X,Y) \in \estDg | (X,Y)\in D^*) \E_{\mathcal D^*} \left[
    \left( Z_1(X) - \E_{\mathcal D^*}[Z_1(X)]  \right) \BR(f(X),Y)|(X,Y)\in \estDg\right] \\
    &+\PP((X,Y) \notin \estDg | (X,Y)\in D^*) \E_{\mathcal D^*} \left[
    \left( Z_1(X) - \E_{\mathcal D^*}[Z_1(X)]  \right) \BR(f(X),Y)|(X,Y)\notin \estDg\right].
\end{align*}
When $D^*$, $\estDg$ and $\tilde D$ have the same feature set, we have
\begin{align*}
    \PP((X,Y)\in D^*|(X,Y) \in \estDg)  = \PP((X,Y) \in \estDg | (X,Y)\in D^*) &= \tau, \\
    \PP((X,Y)\notin D^*|(X,Y) \in \estDg) = \PP((X,Y) \notin \estDg | (X,Y)\in D^*) &= 1-\tau.
\end{align*}
Therefore, 
\[
 \text{Cov}_{\estDisg}(Z_1(X),\BR(f(X),Y)) - \text{Cov}_{\mathcal D^*}(Z_1(X),\BR(f(X),Y)) 
\le 2(1-\tau)(\epsilon_+ + \epsilon_-).
\]
The rest of the proof can be accomplished by following the proof of Theorem~\ref{thm:peerIDN}.
\end{proof}

\section{Proof for Corollaries}

\subsection{Proof for Corollary~\ref{cor:covpeerMul}}\label{proof:covpeerMul}

\begin{proof}

\begin{equation}\label{Eq:peerlossExpSup}
\mathbb E_{ \mathcal{\widetilde D}} [{\ell_{\text{PL}}}(f(X), \widetilde{Y})] = \mathbb E_{\mathcal{\widetilde D}}[\ell(f(X), \widetilde{Y})] -  \mathbb E_{\widetilde{\mathcal{D}}_Y} \left[ \mathbb E_{\mathcal{D}_X}[\ell(f(X_{p}), \widetilde Y_{p})]\right].
\end{equation}
The first term in (\ref{Eq:peerlossExpSup}) is
\begin{align*}
  & \mathbb{E}_\mathcal{\widetilde{D}}[\ell(f(X), \widetilde{Y})] \\
= & \mathbb{E}_\mathcal{{D^*}}\left[ \sum_{j \in [K]} \PP(\widetilde Y = j|X,Y^*) \ell(f(X),j) \right] \\ 
= & \sum_{j \in [K]} \sum_{i\in[K]} \PP(Y^*=i) \mathbb{E}_{\mathcal{D^*}|Y^*=i}[  T_{ij}(X) \ell(f(X),j) ]\\
= & \sum_{j \in [K]} \sum_{i\in[K]} \PP(Y^*=i) \left[   T_{ij} \mathbb{E}_{\mathcal{D^*}|Y^*=i}[\ell(f(X),j) ] +  \text{Cov}_{\mathcal{D^*}|Y^*=i}[  T_{ij}(X), \ell(f(X),j) ]  \right]\\
= & \sum_{j \in [K]} \left[ \mathbb P(Y^*=j) 
\left(1-\sum_{i\ne j,i\in[K]} T_{ji}\right) \mathbb{E}_{\mathcal D^* |{Y^*}=j}\left[ \ell(f(X), j)\right]   + 
\sum_{i\in[K], i\ne j}  \mathbb P(Y^*=i)  T_{ij} \mathbb{E}_{\mathcal D^* |{Y^*}=i} \left[ \ell(f(X), j)  \right] \right] \\
& \qquad + \sum_{j\in[K]}\sum_{i\in[K]} P(Y^* = i)\text{Cov}_{\mathcal D^* |{Y^*}=i} \left[T_{ij}(X), \ell(f(X), j)\right] \\
= & \sum_{j \in [K]} \left[ \mathbb P(Y^*=j) 
\left(1-\sum_{i\ne j,i\in[K]} e_{i}\right) \mathbb{E}_{\mathcal D^* |{Y^*}=j}\left[ \ell(f(X), j)\right]   + 
\sum_{i\in[K], i\ne j}  \mathbb P(Y^*=i)  e_{j} \mathbb{E}_{\mathcal D^* |{Y^*}=i} \left[ \ell(f(X), j)  \right] \right] \\
& \qquad + \sum_{j\in[K]}\sum_{i\in[K]} P(Y^* = i)\text{Cov}_{\mathcal D^* |{Y^*}=i} \left[T_{ij}(X), \ell(f(X), j)\right] \\
= & \left(1-\sum_{i\in[K]} e_{i}\right) 
 \mathbb{E}_{\mathcal D^*}\left[ \ell(f(X), Y^*)\right] + \sum_{j \in [K]} 
\sum_{i\in[K]}  \mathbb P(Y^*=i)  e_{j} \mathbb{E}_{\mathcal D^* |{Y^*}=i} \left[ \ell(f(X), j)  \right] \\
& \qquad + \sum_{j\in[K]}\sum_{i\in[K]} P(Y^* = i)\text{Cov}_{\mathcal D^* |{Y^*}=i} \left[T_{ij}(X), \ell(f(X), j)\right]
\end{align*}

The rest of proofs can be done following standard multi-class peer loss derivations \cite{liu2019peer}. 

\end{proof}

\section{More Discussions}

\subsection{Setting Thresholds $L_{\min}$ and $L_{\max}$}\label{dis:L}

In a high level, there are two strategies for setting $L_{\min}$ and $L_{\max}$: {1) $L_{\min} < L_{\max}$ and 2) $L_{\min} = L_{\max}$}.

\paragraph{Strategy-1: $L_{\min} < L_{\max}$:}
This strategy may provide a higher ratio of true Bayes optimal labels among feasible examples in $\estD$ since some ambiguous examples are dropped. However, dropping examples changes the distribution of $X$ (as well as the distribution of the unobservable $Y^*$), a.k.a. covariate shift \cite{huang2007correcting,cheng2017learningdistill}. 
Importance re-weighting with weight $\gamma(X)$ is necessary for correcting the covariate shift, i.e. the weight of each feasible example $(x,\hat y)\in\estD$ should be changed from $1$ to $\gamma(x)$.
Let $\mathcal D_X$ and $\hat{\mathcal D}_X$ be the marginal distributions of $\mathcal{D}$ and $\estDis$ on $X$.
With a particular kernel $\Phi(X)$, the optimization problem is:
\begin{equation}
\label{eq:opt_gamma}
    \begin{split}
            \min_{\gamma(X)}& \qquad \| \E_{\mathcal D_X} [\Phi(X)] - \E_{\hat{\mathcal D}_{X}} [\gamma(X)\Phi(X)]\| \\
    \text{s.t.} & \qquad\gamma(X) > 0 \text{~~and~~} \E_{\hat{\mathcal D}_{X}}[\gamma(X)] = 1.
    \end{split}
\end{equation}
The optimal solution is supposed to be $\gamma^*(X)=\frac{\PP_{{\mathcal D}_X}(X)}{\PP_{\hat{\mathcal D}_X}(X)}$.
Note the selection of kernel $\Phi(\cdot)$ is non-trivial, especially for complicated features \cite{fang2020rethinking} in DNN solutions.
Using this strategy, with appropriate $L_{\min}$ and $L_{\max}$ such that all the examples in $\estD$ are Bayes optimal, the covariance could be guaranteed to be optimal when each example in $\estD$ is re-weighted by $\gamma^*(X)$.

\paragraph{Strategy-2: $L_{\min} = L_{\max}$:}
Compared with Strategy-1, we effectively lose one degree of freedom for getting a better $\estD$. However, this is not entirely harmful since $\estD$ and $D^*$ have the same feature set, indicating estimating $\gamma(X)$ is no longer necessary and $\gamma(X)=1$ is an optimal solution for (\ref{eq:opt_gamma}) with this strategy.

\paragraph{Strategy selection}
When we can get a high-quality $\estD$ by fine-tuning $L_{\min}$ and $L_{\max}$ or $\estD$ is already provided from other sources, we may solve the optimization problem in (\ref{eq:opt_gamma}) to find the optimal weight $\gamma(X)$.
However, considering the fact that estimating $\gamma(X)$ introduces extra computation and potentially extra errors, we focus on Strategy-2 in this paper.
Using Strategy-2 also reduces the effort on tuning hyperparameters.
Besides, the proposed \covpeer{} loss is tolerant of an imperfect $\estD$ (shown theoretically in Section~\ref{sec:imperfectCov}).

\subsection{Generation of Instance-Dependent Label Noise}\label{sec:instance_noise_gen}

Pseudo codes for generate instance-based label noise are provided in Algorithm \ref{alg_noise}.
This algorithm follows the state-of-the-art method \cite{xia2020parts}.
Define the overall noise rate as $\eta$.

\begin{algorithm}[!h]
\LinesNumbered
\SetAlgoLined
\KwIn{Clean examples ${({x}_{n},y_{n})}_{n=1}^{N}$; Noise rate: $\eta$; Number of classes: $K$; Shape of each feature $x_n$: $S \times 1$.}
Sample instance flip rates $q_n$ from the truncated normal distribution $\mathcal{N}(\eta, 0.1^{2}, [0, 1])$; \hfill \algcom{// mean $\eta$, variance $0.1^2$, range $[0,1]$}\\
Sample   $W \in \mathcal{R}^{S \times K}$ from the standard normal distribution $\mathcal{N}(0,1^{2})$;\\
    \For{$n\in[N]$}{
       $p = {x}_{n}^\top  W$ \hfill \algcom{//  Generate instance dependent flip rates. The size of $p$ is $1\times K$.}\\
       $p_{y_{n}} = -\infty$ \hfill  \algcom{//  Only consider entries that are different from the true label}\\
       $p = q_{n} \cdot \text{softmax}(p) $\hfill \algcom{//  Let $q_n$ be the probability of getting a wrong label}
			\\
	   $p_{y_{n}} = 1 - q_{n}$ \hfill \algcom{//  Keep clean w.p. $1 - q_{n}$} 
			\\
	   Randomly choose a label from the label space as noisy label $\tilde{y}_{n}$ according to $p$;
     } 
 \KwOut{Noisy examples $\{({x}_{i},\tilde{y}_{n}), n\in [N]\}$.}
 \caption{Generating Instance-Dependent Label Noise}
 \label{alg_noise}
\end{algorithm}

Note Algorithm~\ref{alg_noise} cannot ensure $T_{ii}(X)>T_{ij}(X)$ when $\eta>0.5$. To generate an informative dataset, we set $0.9 \cdot T_{ii}(X)$ as the upper bound of $T_{ij}(X)$ and distribute the remaining probability to other classes.

\subsection{Performance without Data Augmentations}\label{sec:noaug}
For a fair comparison with the recent work on instance-dependent label noise \cite{xia2020parts}, we adopt the same data augmentations as \cite{xia2020parts} and re-produce their results using the same noise file as we employed in Table~\ref{table:cifar-inst}.
Each noise rate is tested $5$ times with a different generation matrix $W$ (defined in Algorithm~\ref{alg_noise}).
Table~\ref{table:noaug} shows the advantages of our second-order approach.
\begin{table}[!h]
	\caption{Performance comparisons without data augmentations}
	\begin{center}
	\scalebox{.85}{{
		\begin{tabular}{c|cc} 
			\hline 
			Method & $\eta = 0.2$ & $\eta = 0.4$ \\
			\hline \hline 
			PTD-R-V\cite{xia2020parts} & $69.62 \pm 3.35$ & $64.73 \pm 3.64$ \\
			\covpeer{}  &  $75.52 \pm 3.94$ & $70.30 \pm 2.96$ \\
			\hline 
		\end{tabular}
		}}
	\end{center}
	\label{table:noaug}
\end{table}
\subsection{More Implementation Details on Clothing1M}\label{sec:detailImplement}

\paragraph{Construct $\estD$}
We first train the network for 120 epochs on 1 million noisy training images using the method in \cite{sieve2020}.
The batch-size is set to 32. The initial learning rate is set as 0.01 and reduced by a factor of 10 at 30, 60, 90 epochs. 
We sample 1000 mini-batches from the training data for each epoch while ensuring the (noisy) labels are balanced. 
Mixup \cite{zhang2018mixup} is adopted for data augmentations.
Hyperparameter $\beta$ is set to 0 at first 80 epochs, and linearly increased to 0.4 for next 20 epochs and kept as 0.4 for the rest of the epochs. 
We construct $\estD$ with the best model.

\paragraph{Train with \covpeer{}}
We change the loss to the \covpeer{} loss after getting $\estD$ and continue training the model (without mixup) with an initial learning rate of $10^{-5}$ for 120 epochs (reduced by a factor of 10 at 30, 60, 90 epochs).
We also tested re-train the model with $\estD$ and get an accuracy of $73.56$.
A randomly-collected balanced dataset with $18,976$ noisy examples in each class is employed in training with \covpeer{}.
Examples that are not in this balanced dataset are removed from $\estD$ for ease of implementation.